\pdfoutput=1

\documentclass[11pt]{article}
\usepackage{graphicx}
\usepackage[para]{threeparttable}
\usepackage{array}
\usepackage{wrapfig}
\usepackage{multirow}
\usepackage{amsmath}
\usepackage{amssymb}
\usepackage{pifont}
\usepackage{textcomp}
\usepackage[utf8]{inputenc} 
\usepackage[T1]{fontenc}    
\usepackage{hyperref}       
\usepackage{url}            
\usepackage{booktabs}       
\usepackage{amsfonts}       
\usepackage{nicefrac}       
\usepackage{microtype}      
\usepackage{xcolor}
\definecolor{DeepGreen}{RGB}{34, 139, 34} 
\definecolor{BrownYellow}{RGB}{218, 165, 32} 

\usepackage{ACL2023}

\usepackage{times}
\usepackage{latexsym}
\usepackage[T1]{fontenc}
 
\usepackage[utf8]{inputenc}
\usepackage{microtype}

\usepackage{inconsolata}

\title{Foundations and Recent Trends in Multimodal Mobile Agents: A Survey}

\author{
  \textbf{Biao Wu$^{1*}$, Yanda Li$^{1*}$, Zhiwei Zhang$^{2}$, Yunchao Wei$^{3}$, Meng Fang$^{4}$, Ling Chen$^{1}$} \\
  $^1$Australian Artificial Intelligence Institute, Sydney, Australia\\ 
  $^2$The Pennsylvania State University, Pennsylvania, United States \\
  $^3$Beijing Jiaotong University, Beijing, China \\
  $^4$University of Liverpool, Liverpool, United Kingdom \\
  \texttt{\{biao.wu-2, yanda.li\}@student.uts.edu.au} 
}

\begin{document}

\maketitle


\begin{abstract}
Mobile agents are essential for automating tasks in complex and dynamic mobile environments. As foundation models evolve, they offer increasingly powerful capabilities for understanding and generating natural language, enabling real-time adaptation and processing of multimodal data. This survey provides a comprehensive review of mobile agent technologies, with a focus on recent advancements in foundation models. Our analysis begins by introducing the core components and exploring key representative works in mobile benchmarks and interactive environments, aiming to fully understand the research focuses and their limitations. We then categorize these advancements into two main approaches: prompt-based methods, which utilize large language models (LLMs) for instruction-based task execution, and training-based methods, which fine-tune multimodal models for mobile-specific applications. Finally, unlike existing surveys, we focus on comparing the deployment cost and effectiveness of the two paradigms, summarizing their pros, cons, and suitable application scenarios.



\end{abstract}

\vspace{2mm}
\section{Introduction}
Mobile agents have achieved notable success in handling complex mobile environments, enabling the automation of task execution across various applications with minimal human intervention ~\cite{zhang2023mobile,li2024appagent,bai2024digirl}. These agents are designed to perceive, plan, and execute in dynamic environments, making them highly suitable for mobile platforms that demand real-time adaptability. Over the years, research on mobile agents has evolved from simple rule-based systems to more sophisticated models capable of handling multimodal data and complex decision-making processes ~\cite{shi2017world,rawles2023android}. At the same time, foundation models, such as large language models (LLMs) and multimodal models, have become pivotal in enabling mobile agents to better understand, generate, and adapt to their environments, thus expanding their capabilities in mobile systems ~\cite{ma2024coco,bai2024digirl}.



The growing complexity of mobile environments and the increasing demand for automation highlight the importance of mobile agents ~\cite{deng2024mobile}. They play a critical role in applications such as mobile interfaces, autonomous navigation, and intelligent assistance, allowing for more efficient and intelligent task execution ~\cite{yang2023appagent}. As mobile technologies advance, mobile agents are expected to operate in environments that require continuous adaptation to changing inputs, multimodal data processing, and interaction with various user interfaces ~\cite{zhang2023mobile}. The ability to process and integrate diverse data sources in real-time makes mobile agents essential for enabling seamless user experiences and efficient operations in dynamic mobile platforms.

Despite their progress, mobile agents face several challenges. Traditional evaluation methods often fail to capture real-world mobile tasks' dynamic and interactive nature, limiting their assessment accuracy ~\cite{deng2024mobile}. To address this, recent benchmarks such as AndroidEnv~\cite{toyama2021androidenv} and Mobile-Env~\cite{zhang2023mobile} have been developed to evaluate agents in more realistic, interactive mobile environments, focusing on adaptability and task performance. These benchmarks provide a more comprehensive assessment of mobile agents by measuring task completion and their ability to respond to changes in the environment.

Addressing complex tasks while ensuring mobile agents are multimodal, scalable, adaptable, and resource-efficient remains a significant challenge. Recent advancements in multimodal mobile agent research can be categorized into prompt-based and training-based methods. Prompt-based methods leverage large language models (LLMs), such as ChatGPT~\cite{openai_chatgpt} and GPT-4~\cite{openai2023gpt4}, to handle complex tasks by using instruction prompting and chain-of-thought (CoT) reasoning ~\cite{zhang2024android}. Notable works such as  AppAgent~\cite{yang2023appagent} and AutoDroid~\cite{wen2024autodroid} have demonstrated the potential of prompt-based systems in interactive mobile environments, although scalability and robustness remain ongoing challenges. On the other hand, training-based methods focus on fine-tuning multimodal models, such as LLaVA~\cite{liu2023visual} and Qwen-VL~\cite{bai2023qwen}, specifically for mobile applications. These models can handle rich, multimodal data by integrating visual and textual inputs, improving their ability to perform tasks like interface navigation and task execution ~\cite{ma2024coco,dorka2024training}.

This survey provides a comprehensive review of multimodal mobile agent technologies, focusing on recent advancements and ongoing challenges. First, we explore the core components of mobile agents, including perception, planning, action, and memory, which collectively enable agents to operate effectively in dynamic environments. Furthermore, we explore the benchmarks and evaluation methods used to assess mobile agent performance. Next, we categorize mobile agents into prompt-based and training-based approaches, discussing their strengths and limitations in improving agent adaptability, reasoning, and task execution. Finally, we discuss the future development directions of multimodal mobile agents. By providing a clear understanding of the current state of mobile agent research, this survey identifies key areas for future exploration and offers insights into the development of more adaptive, efficient, and capable mobile agents.

\section{The Components of Mobile Agents}
 
\begin{figure*}[htbp]
    \centering
    \includegraphics[width=0.9\textwidth]{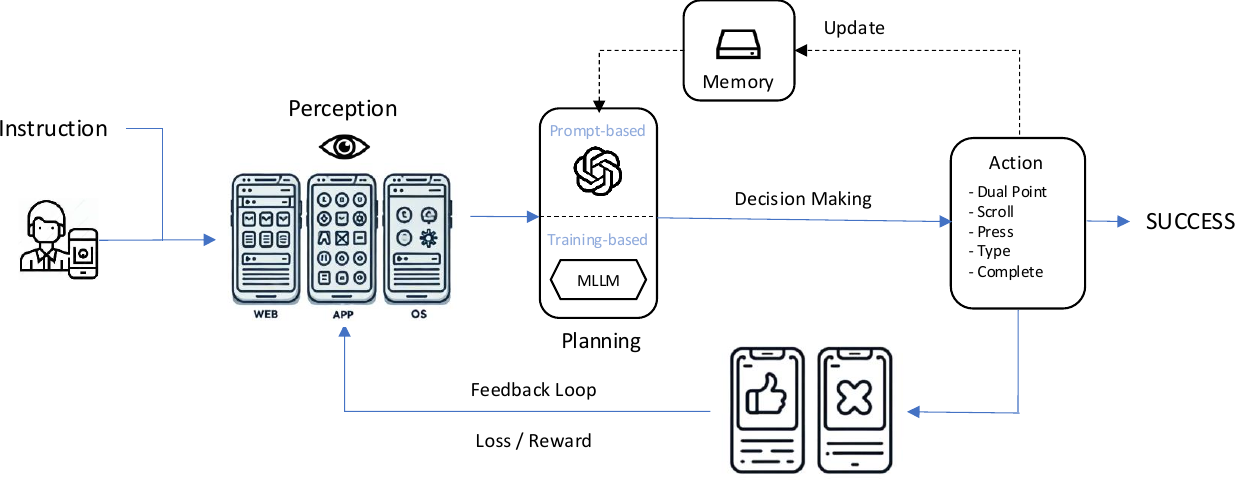}
    \caption{This pipeline shows the decision-making process of mobile agents: User instructions are processed through web, app, and OS interfaces, followed by planning with prompt-based or training-based methods. Actions are taken, and feedback is used to update memory, enabling continuous learning to achieve success.}
    \label{fig:example_image}
\end{figure*}

As shown in Fig. \ref{fig:example_image}, this section outlines the four fundamental components of mobile agents: perception, planning, action, and memory. Together, these components enable agents to perceive, reason, and execute within dynamic mobile environments, adapting their behavior dynamically to improve task efficiency and robustness.

\subsection{Perception}

Perception is the process through which mobile agents gather and interpret multimodal information from their surroundings. In mobile agents, the perception component focuses on handling multimodal information from different environments, extracting relevant information to support planning and task execution. ecent research on perception methods for mobile agents can be broadly categorized into two groups: (1) vision-only approaches, which rely purely on visual inputs, and (2) scene-aware approaches, which incorporate mobile-specific contextual information such as UI structure or APIs.

\paragraph{Vision-Only Approaches.}
Vision-only methods aim to address the generalization challenge in real-world mobile environments, where structured UI representations (e.g., layout trees or metadata) are often inaccessible or unreliable due to encryption, dynamic rendering, or platform constraints. Consequently, researchers turn to vision-centric solutions that rely on raw screen inputs—an approach that mirrors human interaction and offers better cross-platform applicability. Early works~\cite{zhang2021screen,sunkara2022towards,song2023navigating} used simple models to generate textual descriptions from images or audio, but often produced redundant or irrelevant content, impairing downstream planning. Input length limitations in LLMs further exacerbated this issue. To improve visual understanding in mobile contexts, recent works have enhanced the ability of visual encoders to recognize and process key UI elements such as icons and buttons by refining model architectures,  the construction of expert pipelines, designing targeted training strategies, and constructing datasets specifically tailored for mobile interaction scenarios~\cite{cheng2024seeclick,xu-etal-2021-grounding,gou2024navigating,qinghong2024showui,wan2024omniparser,lu2025ui}.

\paragraph{Scene-Aware Approaches.}

In scenarios where structured UI information is available, recent approaches have begun integrating DOM-level cues with functional APIs to enhance both the perception and execution capabilities of autonomous agents. These methods not only parse and rank DOM elements—enabling large language models to more accurately identify and understand interactive components~\cite{li2024appagent,wen2024autodroid,deng2024mobile}—but also leverage APIs to access dynamic information such as the application’s internal state and callable functions. By incorporating structured representations of the interface, these approaches move beyond vision-only inputs and support more accurate and efficient understanding and interaction with mobile UIs~\cite{chai2024amex,li2024effects}. Additionally, Octopus v2~\cite{chen2024octopus} introduces specialized functional tokens to abstract and simplify API usage, significantly improving the efficiency of on-device models while reducing inference latency and computational overhead. The combination of static structure and dynamic interface access provides agents with greater control and scalability in real-world scenarios.

\subsection{Planning}
 
Planning is a core mechanism for mobile agents, enabling them to formulate action strategies in dynamic environments while processing multimodal inputs. Planning strategies are typically categorized as static or dynamic. Static planning decomposes tasks into sub-goals but lacks error correction~\cite{zhang2024android}, whereas dynamic planning adjusts based on real-time feedback, allowing agents to backtrack and re-plan~\cite{gao2023assistgpt, wang2024mobile2}. Recent advances in prompt engineering have further enhanced planning. OmniAct~\cite{kapoor2024omniact} structures multimodal inputs to improve reasoning, enabling agents to dynamically integrate external tools and adapt output formats for more efficient execution.

\subsection{Action}  

The action component demonstrates how agents execute tasks in a mobile environment by utilizing three key aspects: screen interactions, API calls, and agent interactions. Through screen interactions, agents tap, swipe, or type on GUIs, imitating human behavior to navigate apps. They also make API calls to access deeper system functions, such as issuing commands to automate tasks beyond the GUI ~\cite{chen2024octopus}. Additionally, by collaborating with other agents, they enhance their ability to adapt to complex tasks, ensuring efficient task execution across diverse environments ~\cite{zhang2024android}.


\paragraph{Screen Interactions}  
In mobile environments, interactions often involve actions like tapping, swiping, or typing on virtual interfaces. Agents, such as those in AiTW, AITZ, and AMEX~\cite{rawles2024androidinthewild,chai2024amex,zhang2024android}, perform GUI-based actions by mimicking human interactions, ensuring they work smoothly with native apps. These actions go beyond simple gestures, including complex multi-step processes requiring agents to dynamically adapt to changes or new inputs ~\cite{metanlp-2021-meta,wang-etal-2022-motif}.

\paragraph{API Calls}  

Mobile agents rely on various methods to interact with GUIs and perform tasks that require deep integration with mobile operating systems, with API calls serving as the foundation~\cite{chen2024octopus,kapoor2024omniact}. Building on API calls, mobile agents can further leverage HTML and XML data to access core functions, modify device settings, retrieve sensor data, and automate app navigation, extending their capabilities beyond GUI-based inputs~\cite{chai2024amex,chen2024octopus,li2024appagent}. By integrating these approaches, agents can efficiently complete tasks while gaining a more comprehensive understanding of their environment.


\subsection{Memory}

Memory mechanisms are crucial for mobile agents, allowing them to retain and use information across tasks. Current research maps in-context learning to short-term and long-term memory to external vector stores.

\paragraph{Short-term Memory}
Effective task continuity requires short-term memory to retain and reason over recent context. Auto-UI~\cite{zhan2023you} incorporates historical text for better decision-making, while recent works ~\cite{dorka2024training,qin2025ui,qinghong2024showui} stores visual memory. Unlike single-modality agents, multimodal agents must manage short-term memory across text, images, and interactions.

\paragraph{Long-term Memory}
Managing long-term, complex information involves combining parametric memory and vector databases. Parametric memory captures implicit semantics, while vector stores retain recent episodic knowledge. Some approaches convert multimodal inputs into unified text to simplify retrieval and integration~\cite{yang2023appagent,wang2024mobile,wen2024autodroid,li2024appagent,jiang2025appagentx}.

\definecolor{ao}{rgb}{0.0, 0.5, 0.0} 

\begin{table*}[ht]
    \centering
    \small
    \resizebox{1\textwidth}{!}{ 
    
        \begin{tabular}{lllp{2cm}p{3cm}c}
        \toprule
        \textbf{Dataset} & \textbf{Templates} & \textbf{Attach} & \textbf{Task} & \textbf{Reward} & \textbf{Platform} \\ 
        \midrule \textbf{\textit{Static Dataset}} \\

        \texttt{RICOSCA}~\cite{deka2017rico} & 259k & VH & Grounding & - & \textcolor{DeepGreen}{Android} \\
        \texttt{AndroidHowTo}~\cite{deka2017rico} & 10k & - & Extraction & - & \textcolor{DeepGreen}{Android} \\

        \texttt{PixelHelp}~\cite{li2020mapping} & 187 & VH & Apps & - & \textcolor{DeepGreen}{Android} \\  
        \texttt{Screen2Words}~\cite{wang2021screen2words} & 112k & VH & Summarization & - & \textcolor{DeepGreen}{Android} \\
        \texttt{META-GUI}~\cite{sun2022meta} & 1,125 & XML & Apps+Web & - & \textcolor{DeepGreen}{Android} \\
        \texttt{MoTIF}~\cite{burns2021mobile} & 4,707 & VH & Apps & - & \textcolor{DeepGreen}{Android} \\
        \texttt{UGIF}~\cite{venkatesh2022ugif} & 4184 & XML & Grounding & - & \textcolor{DeepGreen}{Android} \\
        \texttt{AitW}~\cite{rawles2024androidinthewild} & 30k & Layout & Apps+Web & - & \textcolor{DeepGreen}{Android} \\
        \texttt{AitZ}~\cite{zhang2024android} & 2504 & CoT & Apps+Web & - & \textcolor{DeepGreen}{Android} \\
        \texttt{AMEX}~\cite{chai2024amex} & 3k & Layout & Apps+Web & - & \textcolor{DeepGreen}{Android} \\

        
        \texttt{Mobile3M}~\cite{chen2024gui} & 3M & XML & Apps & - & \textcolor{DeepGreen}{\textcolor{DeepGreen}{Android}} \\

        \texttt{Androidcontrol}~\cite{li2024effects} & 15283 & VH & Apps+Web & - & \textcolor{DeepGreen}{Android} \\
        
        \texttt{MobileViews-600K}~\cite{gao2024mobileviews} & 600k & VH & Apps & - & \textcolor{DeepGreen}{Android} \\

        \texttt{Ferret-UI}~\cite{you2024ferretuigroundedmobileui} & 120k & VH & Apps & - & \textcolor{black}{IOS} \\
        
        \texttt{Odyssey}~\cite{lu2024gui} & 7735 & - & Apps+Web & - & Multi Platforms \\
        \texttt{ScreenSpot}~\cite{cheng2024seeclick} & 1200 & Layout & Apps+Web & - & Multi Platforms \\
        \texttt{GUI-World}~\cite{chen2024gui} & 12379 & Video & Apps+Web & - & Multi Platforms \\
        
        \midrule \textbf{\textit{Interactive Environment}} \\
        \texttt{MiniWoB++}~\cite{liu2018reinforcement} & 114 & - & Web (synthetic) & Sparse Rewards & - \\
        \texttt{AndroidEnv}~\cite{toyama2021androidenv} & 100 & - & Apps & Sparse Rewards & \textcolor{DeepGreen}{Android} \\
        \texttt{AppBuddy}~\cite{shvo2021appbuddy} & 35 & - & Apps & Sparse Rewards & \textcolor{DeepGreen}{Android} \\
        
        \texttt{Mobile-Env}~\cite{zhang2023mobile} & 224 & VH & Apps+Web & Dense Rewards & \textcolor{DeepGreen}{Android} \\
        
        \texttt{AndroidArena}~\cite{wang2024mobileagentbench} & 221 & XML & Apps+Web & Sparse Rewards & \textcolor{DeepGreen}{Android} \\
        \texttt{AndroidWorld}~\cite{rawles2024androidworld} & 116 & - & Apps+Web & Sparse Rewards & \textcolor{DeepGreen}{Android} \\
        \texttt{DroidTask}~\cite{wen2024autodroid} & 158 & XML & Apps+Web & - & \textcolor{DeepGreen}{Android}  \\
        \texttt{B-MoCA}~\cite{lee2024benchmarking} & 60 & XML & Apps+Web & - & \textcolor{DeepGreen}{Android}  \\

        \texttt{AppWorld}~\cite{trivedi2024appworld} & 750 & API & Apps+Web & - & \textcolor{DeepGreen}{Android}  \\

        \texttt{Mobile-Bench}~\cite{deng2024mobile} & 832 & XML & Apps+Web &  - & \textcolor{DeepGreen}{Android}  \\

        \texttt{MobileAgentBench}~\cite{wang2024mobileagentbench} & 100 & VH & Apps+Web &  Dense Rewards & \textcolor{DeepGreen}{Android}  \\

        \texttt{LlamaTouch}~\cite{zhang2024llamatouch} & 60 & - & Apps+Web &- & \textcolor{DeepGreen}{Android}  \\

        \texttt{Spa-Bench}~\cite{chen2024spa} & 340 & - & Apps+Web & Dense Rewards & \textcolor{DeepGreen}{Android}  \\

        \texttt{AndroidLab}~\cite{xu2024androidlab} & 138 & XML & Apps+Web & Dense Rewards &   \textcolor{DeepGreen}{Android}  \\
        
        \texttt{CRAB}~\cite{xu2024crab} & 23 & XML & Apps+Web & Dense Rewards &   Multi Platforms  \\
      
        \bottomrule
        \end{tabular}
    }
    \caption{Comparison of various platforms based on parallelization, templates, tasks per template, rewards, and supported platforms. Layout refers to fine-grained annotation of the image content, such as the positional coordinates of elements. VH means View Hierarchy. }
    \vspace{-3mm}
    \label{tab:plf_cmp}
\end{table*}



\vspace{3mm}
\section{Mobile Datasets and Benchmarks}

Benchmarks provide standardized environments to evaluate mobile agents on tasks such as UI automation, task completion, and real-world applications. Most existing GUI benchmarks rely on static datasets~\citep{sun2022meta,deng2024mind2web,niu2024screenagent,rossner2020weblinks}, where fixed action sequences serve as ground truth. This rigid evaluation overlooks the diversity of valid strategies and often penalizes correct but non-standard solutions. Interactive benchmarks like AndroidArena~\citep{xing2024understanding} offer more dynamic settings but still rely heavily on action sequence similarity, limiting their ability to assess generalization and reasoning. Recent studies~\citep{yang2023appagent,wang2024mobile2,zhang2024ufo} have explored LLM-based or human evaluation, yet these approaches are typically conducted in uncontrolled environments, lacking reproducibility and consistent evaluation criteria. As a result, researchers are beginning to recognize that improving evaluation metrics alone is insufficient. A more systematic categorization of environments is needed to build a robust benchmarking framework. In the following, we examine three major types of environments—static datasets, simulation environments, and real-world environments—to better understand the current landscape and emerging challenges in benchmarking mobile agents.

\vspace{-1mm}
\subsection{Static Datasets}

Static datasets provide a controlled and predefined set of tasks with annotated ground-truth solutions, making them essential for evaluating mobile agents in fixed environments. They are primarily used to assess task automation, requiring agents to follow specific actions or commands to complete designated tasks. Early work focused on linking referring expressions to UI elements, with each instance containing a screen, a low-level command, and the corresponding UI element. For example, the RicoSCA dataset~\cite{deka2017rico} uses synthetic commands, while MiniWoB++\cite{liu2018reinforcement} includes sequences of low-level actions for multi-step tasks. More recent efforts have shifted toward task-oriented instructions, where each episode consists of action-observation pairs along with screenshots and structured representations such as Android’s View Hierarchy or the web-based Document Object Model. The PixelHelp dataset\cite{li2020mapping} includes 187 high-level task goals with step-by-step instructions from Pixel Phone Help pages, while UGIF~\cite{venkatesh2022ugif} extends similar queries across multiple languages. MoTIF~\cite{burns2021mobile} provides 4.7k task demonstrations with an average of 6.5 steps per task and 276 unique task descriptions. AITW~\cite{rawles2024androidinthewild}, a much larger dataset, includes 715,142 episodes and 30,378 unique prompts, some inspired by prior benchmarks.




\subsection{Simulation Environments}

Simulation environments offer dynamic, real-time interaction platforms that are essential for evaluating agents in complex and evolving scenarios. Unlike static datasets, these environments support continuous adaptation and feedback, making them critical for testing agent flexibility and decision-making. Prior to the emergence of LLM-based agents, research focused on reinforcement learning (RL) systems, such as Android-Env~\cite{toyama2021androidenv}, which relied on predefined actions and rewards. With the advancement of LLMs, attention has shifted toward agents capable of natural language understanding and generation, enabling more flexible, human-like behaviors in tasks like app automation~\cite{wen2023empowering,wen2023droidbot,liu2023chatting,yao2022webshop,shvo2021appbuddy}. Recent efforts, such as Mobile-Env~\cite{zhang2023mobile}, highlight the potential of LLM-based agents to autonomously explore multi-step tasks with minimal reliance on manual scripting, emphasizing adaptability in real-world environments~\cite{liu2024skepticism,sun2024harnessing,sun2024facilitating}.

\subsection{Real-world  Environments}

Real-world  Environments present a significant opportunity to address one of the main limitations of closed-reinforcement learning settings: their inability to fully capture the complexity and variability of real-world interactions. While controlled environments are useful for training and testing agents, they often miss the dynamic elements of real-world scenarios, where factors like changing content, unpredictable user behavior, and diverse device configurations are crucial. To overcome these challenges, researchers are increasingly exploring open, real-world environments for LLM-based GUI agents, enabling them to learn and adapt to the intricacies of live systems and evolving situations~\cite{gao2023assistgui,wang2024mobile,zhang2024ufo,yang2023appagent}. However, deploying agents in open-world settings introduces several risks. These include safety concerns, irreproducible results, and the potential for unfair comparisons. To mitigate these issues and ensure fair, reproducible evaluations, researchers advocate for strategies such as fixing dynamic online content and employing replay mechanisms during evaluation~\cite{liu2018reinforcement,shi2017world,zhou2023webarena}. These methods help create a more controlled testing environment, even within the broader scope of open-world deployments.

\definecolor{ao}{rgb}{0.0, 0.5, 0.0} 
\newcommand{\xmark}{\ding{55}}
\newcommand{\greencheck}{{\color{ao}\checkmark}} 
\newcommand{\redcross}{{\color{red}\xmark}} 

\begin{table*}[t]
    \centering
    \resizebox{1\textwidth}{!}{ 
        \begin{tabular}{l@{\hskip 30pt}c@{\hskip 10pt}c@{\hskip 10pt}c@{\hskip 20pt}c@{\hskip 20pt}c}
        \toprule
        \textbf{Method} & \textbf{Input Type} & \textbf{Model} & \textbf{Training} & \textbf{Memory} & \textbf{Multi-agents} \\ 
        
        \midrule \textbf{\textit{Prompt-based Methods}} \\ 
        DroidGPT~\cite{wen2023droidbot} & Text & ChatGPT & None & \redcross & \redcross \\
        AppAgent~\cite{yang2023appagent} & Image\&Text & GPT-4V & None & \greencheck & \redcross \\
        
        MobileAgent~\cite{wang2024mobile} & Image\&Text & GPT-4V \& DINO \& OCR & None & \greencheck & \redcross \\
        MobileAgent v2~\cite{wang2024mobile2} & Image\&Text & GPT-4V \& DINO \& OCR & None & \greencheck & \greencheck \\


        AutoDroid~\cite{wen2024autodroid} & Text & GPT-4 \& Vicuna-7B & None & \greencheck & \greencheck \\
        AppAgent V2~\cite{li2024appagent} & Image\&Text & GPT-4V  & None & \greencheck & \redcross \\
        VLUI~\cite{lee2024benchmarking} & Image\&Text & GPT-4V & None & \redcross &\redcross \\

        MobileExperts~\cite{zhang2024mobileexperts} & Image\&Text & GPT-4V & None & \redcross &  \greencheck \\
        Mobile-Agent-E~\cite{wang2025mobile} & Image\&Text & GPT-4o \& DINO \& Qwen-VL-Plus  & None & \greencheck & \greencheck \\
        \midrule \textbf{\textit{Training-based Methods}} \\
        MiniWob~\cite{liu2018reinforcement} & Image & DOMNET & RL-based & \redcross & \redcross \\
        MetaGUI~\cite{sun2022meta} & Image\&Text & Faster-RCNN \& Transformer & Pre-trained & \redcross & \redcross \\
        CogAgent~\cite{hong2023cogagent} & Image\&Text & CogVLM & Pre-trained & \redcross & \redcross \\
        SeeClick~\cite{cheng2024seeclick} & Image\&Text & Qwen-VL & Pre-trained & \redcross & \redcross \\

        AutoGUI~\cite{zhang2023you} & Image\&Text & BLIP-2-ViT \& FLAN-Alpaca  & Finetune & \greencheck & \redcross \\
        ResponsibleTA~\cite{zhang2023responsible} & Image\&Text & Swin-ViT \& BART & Finetune & \greencheck & \redcross \\
        
        UI-VLM~\cite{dorka2024training} & Image\&Text & Qwen-VL & Finetune & \greencheck & \redcross \\
        Coco-Agent~\cite{ma2024coco} & Image\&Text & CLIP-ViT \& LLama2-7B & Finetune & \greencheck & \redcross \\
        DigiRL~\cite{bai2024digirl} & Image\&Text & BLIP-2-ViT \& Flan-T5 \& RoBerta & RL-based & \redcross & \redcross \\
        SphAgent~\cite{chai2024amex} & Image\&Text & SPHINX-X & Finetune & \redcross & \redcross \\
        Octopus v2~\cite{chen2024octopus}  & Text & Gemma-2B & Finetune & \redcross & \redcross \\
        Octo-planner~\cite{chen2024octo}  & Text & Phi-3 Mini & Finetune & \redcross & \greencheck \\
        MobileVLM~\cite{wu2024mobilevlm} & Image\&Text & Qwen-VL-Chat & Finetune & \redcross &\redcross \\
        OdysseyAgent ~\cite{lu2024gui} & Image\&Text & Qwen-VL & Finetune & \greencheck &\redcross \\

        AutoGLM ~\cite{liu2024autoglm} & Image\&Text & ChatGLM & RL-based & \redcross&\redcross \\
        LiMAC ~\cite{christianos2024lightweight} & Image\&Text & Qwen2-VL-2B \& AcT & Finetune & \redcross&\redcross \\
        DistRL ~\cite{wang2024distrl} & Image\&Text & BLIP-2-ViT \& Flan-T5 & RL-based & \redcross&\redcross \\
        ShowUI~\cite{qinghong2024showui} & Image\&Text & Qwen2-VL-2B & Finetune & \redcross & \redcross \\
        OmniParser~\cite{wan2024omniparser} & Image\&Text & BLIP-2 \& OCR \& YoloV8 & Finetune & \redcross & \redcross \\
        OS-Atlas ~\cite{wu2024atlas} & Image\&Text & InternVL-2-4B \textbackslash{}  Qwen2-VL-7B & Pre-trained & \redcross & \redcross \\
        AppVLM ~\cite{papoudakis2025appvlm} & Image\&Text & Paligemma-3B-896 & RL-based & \redcross & \redcross \\
        InfiGUIAgent ~\cite{liu2025infiguiagent} & Image\&Text & Qwen2-VL-2B & Finetune & \redcross & \redcross \\
        UI-Atlas ~\cite{qin2025ui} & Image\&Text & Qwen2-VL-72B & Finetune & \redcross & \redcross \\
        \bottomrule
        \end{tabular}
        }
    \caption{Comparison of Mobile Agents: A Detailed Overview of Input Types, Models, Training Methods, Memory Capabilities, and Multi-agent Support.}
    \label{tab:mobile_agent}
\vspace{-3mm}
\end{table*}

\subsection{Evaluation Methods}

In evaluating agent performance, trajectory evaluation, and outcome evaluation are two main methods. Trajectory evaluation focuses on how well agent actions align with predefined paths. In contrast, outcome evaluation emphasizes whether the agent achieves its final goals, focusing on results rather than the specific process. The following sections will explore recent research advancements in these two areas, highlighting how more comprehensive evaluation strategies can enhance our understanding of agent performance in complex environments.

\paragraph{Trajectory Evaluation}

Recent improvements in GUI interaction benchmarks have focused on step-by-step assessments, comparing predicted actions to reference action trajectories to evaluate agent performance effectiveness ~\cite{rawles2024androidinthewild,zhang2021screen}. While this approach is effective in many cases, task completion often has multiple valid solutions, and agents might explore different paths that do not necessarily follow the predefined trajectories. To improve the flexibility and robustness of these evaluations, Mobile-Env evaluate a subset of signals from the environment of an intermediate state, enabling reliable assessment across a wider range of tasks~\cite{zhang2023mobile}.

\paragraph{Outcome Evaluation}

An agent's success is determined by assessing whether it reaches the desired final state, treating task goals as subsets of hidden states, regardless of the path taken to achieve them. These final states can be identified through various system signals. Relying on a single signal type may not capture all relevant state transitions, as certain actions, such as form submissions, may only be visible in the GUI and not in system logs \cite{toyama2021androidenv} or databases \cite{rawles2024androidworld}.  Shifting to outcome-based evaluation and using multiple signals can make GUI interaction benchmarks more reliable and adaptable, allowing agents to show their full abilities in various scenarios ~\cite{wang2024mobileagentbench,rawles2024androidworld}.

\subsection{Performance Comparison}

Due to the limitations of current benchmarks, variations in implementation methods, and changes in platforms, comparing all methods within a unified evaluation environment is challenging.  Meanwhile, both prompt-based and training-based approaches suffer from inconsistent evaluation metrics, complicating cross-study comparisons. Methods such as AppAgent \cite{li2024appagent} and AutoDroid \cite{wen2024autodroid} introduce their own benchmarks and metrics, but only test within these benchmarks and compare against models like GPT-4. These disparities make direct experimental comparisons impractical at this stage.  Therefore, after reviewing experimental results from different studies, we compared the AITW and MobileAgentbench benchmarks. AITW measures instruction accuracy \cite{rawles2024androidworld}, and MobileAgentbench measures Success Rate \cite{wang2024mobileagentbench}. See tables \ref{table:part3} and \ref{tab:experimental_results_mobile_agents} in the appendix for more details and the need for standardized benchmarks in future research.



\definecolor{ao}{rgb}{0.0, 0.5, 0.0} 


\section{The Taxonomy of Mobile Agents}

This section introduces a taxonomy of mobile agents, categorizing them into two primary types: prompt-based methods and training-based methods. As shown in Table ~\ref{tab:mobile_agent}, prompt-based agents take advantage of advancements in LLM to interpret and execute instructions through natural language processing, often focusing on tasks that require dynamic interaction with GUI. Training-based methods involve fine-tuning models or applying reinforcement learning to enhance agents' decision-making and adaptability over time.

\subsection{Prompt-based Methods}

Recent advancements in LLMs have demonstrated significant potential in developing autonomous GUI agents, particularly in tasks that require instruction following \cite{sanh2021multitask,taori2023stanford,chiang2023vicuna} and chain-of-thought (CoT) prompting \cite{nye2022show,cot_wei}. CoT prompting \cite{cot_wei,kojima2022large,zhang2022automatic}, in particular, has proven effective in enabling LLMs to handle step-by-step processes, make decisions, and execute actions. These capabilities have shown to be highly beneficial in tasks involving GUI control \cite{rawles2023android}.


\paragraph{Perception Tools} Enabling LLMs to interact with GUI is essential, as these models are primarily designed to process natural language rather than visual elements. Perception tools play a crucial role in bridging this gap by allowing LLMs to interpret and interact with visual elements through text-based commands, making it possible for the models to process and respond to graphical interface components. This multimodal integration significantly boosts the efficiency and flexibility of mobile agents in complex environments.
Techniques like icon recognition and OCR \citep{zhang2021screen,sunkara2022towards,song2023navigating} are used to parse GUI elements, which then converts the parsed elements into HTML layouts. However, this method relies heavily on external tools \citep{rawles2023android,wen2023empowering} and app-specific APIs \citep{zhou2023webarena,gur2023real}, often resulting in inefficiencies and errors during inference. Although some research has investigated multimodal architectures to process different types of inputs \citep{sun2022meta,yan2023gpt}, these approaches still depend on detailed environment parsing for optimal performance. Given the importance of accurate GUI grounding, newer studies \citep{cheng2024seeclick,hong2023cogagent,gou2024navigating,qinghong2024showui} have begun exploring pre-training methods to improve agent performance in GUI tasks.



\paragraph{Memory Mechanism} Effective task execution in prompt-based methods relies on a strong memory mechanism to retain and use relevant information.  In agents like AppAgent ~\cite{yang2023appagent}, the agent employs an exploration phase for memory, allowing it to learn and adapt to new applications by storing interactions from prior explorations. This approach enables the agent to retain knowledge without needing additional training data. Mobile-Agent ~\cite{wang2024mobile,wang2024mobile2} automates mobile app operations by analyzing screenshots with perception tools, avoiding reliance on system code. 





\subsection{Training-based Methods}

In contrast to prompt-based methods, training-based approaches involve explicit model optimization. These agents fine-tune large language models like LLama~\cite{zhang2023llama} or multimodal models such as LLaVA~\cite{liu2023visual} by collecting instruction-following data to obtain instruction information. 



\paragraph{Pre-trained VLMs} In mobile environments, pre-trained VLMs have become powerful tools for decision-making and interaction. Models like LLaVA~\cite{liu2023visual} and Qwen-VL~\cite{bai2023qwen}, pre-trained on large-scale general datasets, capture both visual and language information effectively. However, their applicability in mobile settings is limited by the lack of sensitivity to interactive elements specific to mobile data. To improve the responsiveness of pre-trained models to interactive elements in mobile data, CogAgent~\cite{hong2023cogagent} collected a large-scale mobile dataset for pre-training representations. CogAgent~\cite{hong2023cogagent} integrates visual and textual inputs for GUI agents, improving interaction with complex mobile UIs using VLMs. Spotlight~\cite{li2022spotlight} is a vision-language model for mobile UI tasks, relying solely on screenshots and specific regions, supporting multi-task and few-shot learning, trained on a large-scale dataset. VUT~\cite{li2021vut} employs a dual-tower Transformer for multi-task UI modeling, achieving competitive performance with fewer models and reduced computational costs.


\paragraph{Fine-Tuning}

The process of fine-tuning pre-trained VLMs with commonsense reasoning capabilities has been facilitated by large-scale mobile datasets, such as AitW~\cite{rawles2024androidinthewild}, through the Visual Instruction Tuning approach. Existing methods primarily involve two areas: dataset enhancement, and training strategies improvement. ScreenAI~\cite{baechler2024screenai} and AMEX~\cite{chai2024amex} focus on using synthetic data and multi-level annotations to precisely identify and describe UI elements on mobile interfaces, providing high-quality datasets for complex question-answering and navigation tasks. On the other hand, Auto-GUI~\cite{zhan2023you}, UI-VLM~\cite{dorka2024training}, COCO-Agent~\cite{ma2024coco}, Octo-planner~\cite{chen2024octo}, and AutoDroid~\cite{wen2024autodroid} achieve significant model performance improvements through strategies such as direct interface interaction, task instruction and element layout improvement, and separating planning from execution. These techniques not only optimize automation processes but also enhance the prediction accuracy and operational efficiency of models in practical applications.



\paragraph{Reinforcement Learning} 
A dynamic approach to training mobile agents is offered by reinforcement learning, which enables them to learn from interactions with environments. This method is particularly effective in scenarios where the agent must adapt to sequential decision-making tasks or optimize its actions based on rewards. The WoB~\cite{shi2017world} platform enables reinforcement learning in real environments by allowing agents to interact with websites using human-like actions. Meanwhile ~\cite{shi2017world} converts action prediction into question-answering, improving task generalization across different environments. MiniWoB++~\cite{liu2018reinforcement} introduces workflow-guided exploration, which integrates expert workflows with task-specific actions, accelerating learning and improving task efficiency in action prediction tasks. DigiRL~\cite{bai2024digirl} combines offline and online reinforcement learning to train device control agents. It scales online training using a VLM-based evaluator that supports real-time interaction with 64 Android emulators, enhancing the efficiency of RL-based agent training.

\section{Trade-off Analysis}

To evaluate the practical trade-offs between prompt-based and training-based approaches, we conducted a series of real-world app automation tasks. These included automatically monitoring price information on e-commerce pages, aggregating and summarizing product data, analyzing cost-effectiveness relative to user preferences, and completing the checkout process. Our observations reveal clear distinctions in deployment cost, inference speed, operational expense, and data privacy between the two paradigms.

Prompt-based agents rely on commercial APIs such as GPT-4V and Gemini, requiring no local deployment and enabling rapid prototyping with minimal maintenance. However, each task incurs a relatively high cost (approximately $0.70–$1.20), with slower inference speeds (5–25 seconds per step), and all user data is processed externally—raising privacy concerns. In contrast, training-based agents require upfront access to dedicated infrastructure, such as a server equipped with two A100 GPUs. While the purchase cost is approximately \$30,000, we opted for a rental-based setup during our 56-day evaluation, with a total rental cost of around \$4,000. Once deployed, training-based agents offer significantly lower per-task costs (\$0.01–\$0.05), faster inference (1–3 seconds per step), and full control over data privacy. However, this approach requires more engineering effort for deployment, fine-tuning, and ongoing system maintenance.

\begin{table}[t]
\centering
\resizebox{0.48\textwidth}{!}{
\begin{tabular}{lcc}
\toprule
\textbf{Aspect} & \textbf{Prompt-based} & \textbf{Training-based} \\
\midrule
Model & GPT4o   & Qwen-VL-72B \\
Deployment & -   & 2xA100 \\
Hardware Cost & - &   \$4000 \\
Running Cost  & \$0.70 -- \$1.20  & \$0.01 -- \$0.05   \\
Inference Speed & 5--25s  & 1--3s  \\
Data Privacy & Low   & High   \\
Flexibility & High   & Medium   \\
Maintenance Effort & Minimal & Medium to High \\
Success Rate & 72\% & 43\% \\
Cost per Completion &  \$0.87   & \$0.02  \\

\bottomrule
\end{tabular}
}
\caption{Estimated comparison between Prompt-based and Training-based mobile agents. Hardware cost reflects the estimated price of a 2×A100 server setup. Running cost is calculated per task execution, while inference speed refers to the average time per step. Estimates are based on GPT-4V API usage and local deployment of a 72B model.}
\label{tab:prompt_vs_train_estimate}
\vspace{-3mm}
\end{table}

Overall, prompt-based solutions are more suitable for lightweight or rapidly changing tasks, while training-based agents offer superior long-term efficiency, privacy, and scalability for high-frequency or latency-sensitive applications.

\section{Conclusion}


This survey provides a comprehensive overview of multimodal mobile agent technologies. Firstly,  we discussed the core components—perception, planning, action, and memory—that enable mobile agents to adapt to their environments, forming the foundation of their functionality. Next, we reviewed advancements in mobile agents' benchmarks, which improve mobile agent assessments but still require more comprehensive methods to capture real-world dynamics. We then presented a taxonomy of mobile agents, differentiating between prompt-based and training-based methods, each with strengths and challenges in scalability and adaptability.
Finally, we highlighted future research directions, focusing on security, adaptability, and multi-agent collaboration to advance mobile agent capabilities.

\section{Limitations}

This survey focuses on recent advancements in LLM-based mobile agents but provides limited coverage of traditional, non-LLM-based systems. The lack of discussion on older rule-based agents may limit the broader context of mobile agent technology development.

\bibliography{anthology,custom}
\bibliographystyle{acl_natbib}
\clearpage
\appendix

\section{ Appendix}

\subsection{Future Work}

In this survey, we have presented the latest advancements in the field of mobile agents. While significant progress has been made, several challenges remain unresolved. Based on the current state of research, we propose the following future research directions:

\paragraph{Model Architecture Optimization :}
When optimizing mobile agent performance, it is crucial to consider the impact of grounding ability on the action prediction task. To achieve this, the model must enhance its grounding ability for accurate element localization while effectively adapting to action prediction tasks and making efficient decisions. The Mixture of Experts (MOE) architecture plays a key role in this process~\cite{lin2024moe}. By introducing multiple expert modules, MOE allows the model to dynamically select the most suitable expert based on the task, making it particularly effective in handling multi-domain tasks and improving task adaptability and performance~\cite{shu2024llava}. Therefore, adopting the MOE architecture enhances grounding ability while ensuring strong decision-making capability in complex tasks, leading to improved performance across multiple domains~\cite{wu2024deepseek}.

\paragraph{Combine with Reinforcement Learning:} Enhancing mobile agents' ability to adapt to dynamic and unpredictable environments is crucial. Mobile agent tasks are fundamentally decision-making tasks, not just prediction tasks \cite{liu2018reinforcement}. Training through instruction fine-tuning can improve predictions within the action space, but it struggles with decision data influenced by predicted outcomes that cause distribution changes, such as in virtual machines or simulators. These scenarios require the use of reinforcement learning to perform sequential decision-making tasks. However, research in this area is still in its early stages. Current explorations, such as Digirl \cite{bai2024digirl}, Distrl\cite{wang2024distrl} and RL4VLM \cite{zhai2024fine}, have not yet achieved end-to-end alignment in this field. Future research should explore how to utilize reinforcement learning better to integrate changing interactive environments with multimodal large language models for real-time behavior adjustment.

\paragraph{Security and Privacy:} 
Mobile agents face security risks in open environments. Whether it involves tasks that make decisions in latent spaces, such as those found in the AITW \cite{rawles2024androidworld} and AMEX \cite{chai2024amex} datasets, or tasks like AITZ \cite{zhang2024android} that complete decision-making through chains-of-thought , the security of the model and its ethical aspects can impact decision-making performance \cite{bai2022training} . Future work should prioritize stronger security mechanisms to prevent malicious behavior and data breaches. It’s also necessary to develop privacy protection technologies and ethical improvement mechanisms to ensure safe and ethical operations during agent interactions.

\paragraph{Multi-agent Collaboration:} Collective intelligence simplifies complex problems through distributed control, enhances system robustness with redundant designs, and optimizes resource usage through coordinated operations, thereby demonstrating significant efficiency and adaptability in handling large-scale, complex tasks. Improving collaboration among multiple mobile agents remains a key challenge. Current methods exploring multi-agent systems are still limited to role-playing\cite{li2024appagent}, standard operating procedures \cite{zhang2024mobileexperts}, and collaboration with expert models \cite{chen2024octo}. The overall scale is small, lacking exploration into communication and organizational structures. Future research should focus on efficient communication and collaborative mechanisms that enable agents to dynamically form coalitions and complete tasks more effectively.

\paragraph{Model Lightweighting:} The computational resources on mobile devices are limited, which imposes higher requirements for model deployment and inference. Therefore, quantization and accelerating inference are particularly important. Existing methods such as SphAgent \cite{chai2024amex}, CogAgent \cite{hong2023cogagent}, and SeeClick \cite{cheng2024seeclick} still have too large parameter sizes for deployment on mobile devices. The latest research, like LiMAC \cite{christianos2024lightweight}, reduces fine-tuning costs without compressing model parameters. Future research should focus on optimizing the size of mobile agents and speeding up the inference process to ensure high performance under resource constraints. Additionally, refining the inference pipeline to enhance real-time decision-making capabilities is crucial, which involves better computational algorithms and hardware accelerations to achieve faster responses and reduce energy consumption.

\begin{table*}[htb]
	\centering\small 
	\begin{tabular}{p{5.8cm}p{1.1cm}p{1.1cm}p{1.1cm}p{1.8cm}p{1.1cm}p{1.1cm}p{1.1cm}}
		\toprule
		\textbf{Model} &\textbf{Overall} &\textbf{General} & \textbf{Install} & \textbf{GoogleApps} &\textbf{Single} &\textbf{WebShop.} \\
        \midrule
       
        ChatGPT-COT \cite{ding2024mobileagent} &7.72& 5.93 &4.38 &10.47& 9.39& 8.42\\
        GPT-4V ZS+HTML \cite{ding2024mobileagent}  & 50.54 & 41.66 & 42.64 & 49.82 & 72.83 & 45.73 \\
        GPT-4V ZS+History \cite{ding2024mobileagent}  & 52.96 & 43.01 & 46.14 & 49.18 & 78.29 & 48.18 \\
        GPT-4o \cite{wu2024mobilevlm} & 55.02 & 47.06 & 49.12 & 52.30 & 80.28 & 46.42 \\
        MobileAgent \cite{wang2024mobile} & 66.92 & 55.8 & 74.98 & 63.95 & 76.27 & 63.61 \\
         \midrule

        InternVL +History \cite{wu2024mobilevlm} & 2.63 & 1.95 & 2.88 & 2.94 & 3.03 & 2.71 \\
        Qwen-VL +History \cite{wu2024mobilevlm} & 3.23 & 2.71 & 4.11 & 4.02 & 3.89 & 2.58 \\
        PaLM-2 \cite{zhang2023you} & 39.6 & -- & -- & -- & -- & --  \\
        
        MM-Navigator \cite{yan2023gpt} & 50.54  &41.66  &42.64  &49.82  &72.83  &45.73 \\
        MM-Navigator$_{\textup{w/ text}}$ \cite{yan2023gpt} & 51.92 &42.44  &49.18  &48.26  &76.34  &43.35 \\
        MM-Navigator$_{\textup{w/ history}}$ \cite{yan2023gpt} & 52.96  &43.01  &46.14  &49.18  &78.29  &48.18\\

        OmniParser \cite{wan2024omniparser} & 50.54  &41.66  &42.64  &49.82  &72.83  &45.73 \\
       
        \midrule
        
        BC \cite{rawles2023android} & 68.7 & -- & -- & -- & -- & -- \\
        BC $_{\textup{w/ history}}$ \cite{rawles2023android}  & 73.1 &63.7 &77.5 &75.7 &80.3 &68.5\\  

        Qwen-2-VL \cite{wang2024qwen2} & 67.20 & 61.40 & 71.80 & 62.60 & 73.70 & 66.70 \\
       
        Show-UI \cite{qinghong2024showui} & 70.00 & 63.90 & 72.50 & 69.70 & 77.50 &  66.60 \\

        Llama 2 \cite{zhang2023you} & 28.40 & 28.56 & 35.18 & 30.99 & 27.35 & 19.92 \\
        Llama 2+Plan+Hist \cite{zhang2023you} & 62.86 & 53.77 & 69.1 & 61.19 & 73.51 & 56.74 \\
        
        Auto-UI \cite{zhang2023you} & 74.27 &68.24 &76.89 &71.37 &84.58 &70.26 \\
        MobileVLM \cite{wu2024mobilevlm} & 74.94 & 69.58 & 79.87 & 74.72 & 81.24 & 71.70 \\
        SphAgent \cite{chai2024amex}  & 76.28  &68.20 & 80.50 & 73.30 & 85.40 & 74.00 \\

        CoCo-LLAVA \cite{ma2024coco} & 70.37 & 58.93 & 72.41 & 70.81 & 83.73 & 65.98 \\
        
        SeeClick \cite{cheng2024seeclick} & 76.20 & 67.60 &  79.60 & 75.90 & 84.60 & 73.10 \\
        CogAgent \cite{hong2023cogagent}  & 76.88 & 65.38 & 78.86 & 74.95 & 93.49 & 71.73 \\    
		\bottomrule
	\end{tabular}
	\caption{Experimental results of different methods on static dataset AITW: Action accuracy across main setups, highlighting overall performance in decision-making tasks. The symbol \textasteriskcentered{} indicates that CoCo-Agent incorporates layout information during training, while other models rely solely on image data. }
	\label{table:part3}
\end{table*}

\begin{table*}[h]
    \centering
    \resizebox{0.8\textwidth}{!}{
    \renewcommand{\arraystretch}{1.2}
        \begin{tabular}{lcccccccccc}
        \toprule
        \multirow{2}{*}{\textbf{Model}} & \multirow{2}{*}{Method} & Model & Data & \multicolumn{2}{c}{Mobile} & \multicolumn{2}{c}{Web} & \multicolumn{2}{c}{Desktop} & \multirow{2}{*}{Average} \\
        \cmidrule(lr){5-6} \cmidrule(lr){7-8} \cmidrule(lr){9-10}
         & & Size & Size & Icon & Text & Icon & Text & Icon & Text &  \\
        \midrule
        \multicolumn{11}{l}{\textbf{Supervised Fine-tuning}} \\
        \midrule
        CogAgent   & SFT & 18B  & -    & 24.0 & 67.0 & 28.6 & 70.4 & 20.0 & 74.2 & 47.4 \\
        SeeClick   & SFT & 9.6B & 1M   & 52.0 & 78.0 & 32.5 & 55.7 & 30.0 & 72.2 & 53.4 \\
        UGround-V1 & SFT & 7B   & 10M  & 60.3 & 82.8 & 70.4 & 80.4 & 63.6 & 82.5 & 73.3 \\
        Qwen2.5-VL & SFT & 3B   & 500  & 71.2 & 95.2 & 63.1 & 78.3 & 46.4 & 85.0 & 75.7 \\
        AGUVIS     & SFT & 7B   & 1M   & 78.2 & 88.3 & 70.7 & \underline{88.1} & \underline{74.8} & 85.7 & 81.8 \\
        \midrule
        \multicolumn{11}{l}{\textbf{Zero Shot / Reinforcement Learning}} \\
        \midrule
        Qwen2-VL   & ZS  & 7B   & 0    & 60.7 & 75.5 & 25.7 & 35.2 & 54.3 & 76.3 & 55.3 \\
        Qwen2.5-VL & ZS  & 3B   & 0    & 61.1 & 90.5 & 43.2 & 60.0 & 40.0 & 80.9 & 65.0 \\
        UI-R1-3B   & RFT & 3B   & 136  & \textbf{84.7} & \underline{95.6} & \underline{73.3} & 85.2 & 59.3 & \underline{90.2} & \underline{83.3} \\
        \bottomrule
    \end{tabular}
    }
    \caption{Grounding accuracy on ScreenSpot. The optimal and the suboptimal results are \textbf{bolded} and \underline{underlined}, respectively. ZS indicates zero-shot OOD inference and RFT indicates rule-based reinforcement learning.}
    \label{tab:ss}
\end{table*}


\begin{table*}[htbp]
\centering
\resizebox{0.8\textwidth}{!}{
\begin{tabular}{lccc}
\toprule
\textbf{Model} & \textbf{Model Size} & \textbf{Screen Representation} & \textbf{Success Rate (pass@1)} \\
\midrule
V-Droid (Llama8B)        & 8B   & A11y tree    & 59.5 \\
Agent S2                 & -    & Screenshot   & 54.3 \\
UI-TARS                  & 72B  & Screenshot   & 46.6 \\
GPT-4o + Aria-UI         & -    & Screenshot   & 44.8 \\
GPT-4o + UGround         & -    & Screenshot   & 44.0 \\
GPT-4o (Ponder \& Press) & -    & Screenshot   & 34.5 \\
GPT-4 Turbo              & -    & A11y tree    & 30.6 \\
Qwen2-VL-2B (fine-tuned) & 2B   & Screenshot   & 9.0  \\
\bottomrule
\end{tabular}
}
\caption{Leaderboard of mobile agents on AndroidWorld, comparing model type, screen representation, and pass@1 success rate.}
\label{tab:mobile_agent_comparison_slim}
\end{table*}

\begin{table*}[ht]
    \centering
    \resizebox{0.95\textwidth}{!}{%
        \begin{tabular}{lcccccc}
        \toprule
        \textbf{Agent} & \textbf{SR $\uparrow$} & \textbf{SE $\downarrow$} & \textbf{Latency (s) $\downarrow$} & \textbf{Tokens $\downarrow$} & \textbf{FN Rate $\downarrow$} & \textbf{FP Rate $\downarrow$} \\ 
        \midrule 
        AndroidArena~\cite{xing2024understanding} & 0.22 & 1.13 & 18.61 & 750.47 & 0.09 & 0.33 \\
        AutoDroid~\cite{wen2024autodroid} & 0.27 & 3.10 & 4.85 & 963.48 & 0.93 & 0.01 \\
        AppAgent~\cite{yang2023appagent} & 0.40 & 1.29 & 26.09 & 1505.09 & 0.17 & 0.40 \\
        CogAgent~\cite{hong2023cogagent} & 0.08 & 2.42 & 6.76 & 579.84 & 1.00 & 0.04 \\
        MobileAgent~\cite{ding2024mobileagent} & 0.26 & 1.33 & 15.91 & 1236.88 & 0.19 & 0.31 \\
        \bottomrule
        \end{tabular}
    }
    \caption{Experimental results of different methods on simulation environment MobileAgentBench. SR (Success Rate), SE (System Error), Latency, Tokens, FN Rate (False Negative Rate), and FP Rate (False Positive Rate) are the metrics used for comparison.}
    \label{tab:experimental_results_mobile_agents}
    \vspace{5mm}
\end{table*}

\subsection{Complementary Technologies}


Effective complementary technologies are vital for enhancing the performance and usability of mobile agents, in addition to key components like benchmarks, VLM models, fine-tuning methods, and advanced reasoning skills. These technologies facilitate seamless interactions with mobile environments, allowing agents to adapt, learn, and perform complex tasks efficiently. 


UIED~\cite{xie2020uied} detects and classifies GUI elements using computer vision and deep learning, supporting interactive editing. WebGPT~\cite{nakano2021webgpt} fine-tunes GPT-3 for web-based question answering using imitation learning and human feedback. WebVLN~\cite{chen2024webvln} trains AI agents to navigate websites with question-based instructions, incorporating HTML for deeper understanding.

\subsection{Available related technologies}

\begin{table*}[ht]
    \centering
    \small
    \resizebox{1\textwidth}{!}{ 
    
        \begin{tabular}{lllp{2cm}p{3cm}c}
        \toprule
        \textbf{Dataset} & \textbf{Templates} & \textbf{Attach} & \textbf{Task} & \textbf{Reward} & \textbf{Platform} \\ 
        \midrule \textbf{\textit{Static Dataset}} \\


        \texttt{WebSRC}~\cite{chen2021websrc} & 400k & HTML & Web & - & \textcolor{blue}{Windows}  \\
        \texttt{WebUI}~\cite{wu2023webui} & 400k & HTML & Web & - & \textcolor{blue}{Windows}  \\
        \texttt{Mind2Web}~\cite{deng2024mind2web} & 2,350 & HTML & Web & - & \textcolor{blue}{Windows}  \\
        \texttt{Ferret-UI}~\cite{you2024ferretuigroundedmobileui} & 120k & - & Apps & - & \textcolor{black}{IOS} \\

        \texttt{OmniAct}~\cite{kapoor2024omniact} & 9802 & Ocr/Seg & Web & - & \textcolor{blue}{Windows} \\
        \texttt{WebLINX}~\cite{rossner2020weblinks} & 2,337 & HTML & Web & - & \textcolor{blue}{Windows}  \\
        \texttt{ScreenAgent}~\cite{niu2024screenagent} & 3005 & HTML & Web & - & \textcolor{blue}{Windows}  \\

        \midrule \textbf{\textit{Interactive Environment}} \\
        \texttt{WebShop}~\cite{yao2022webshop} & 12k & - & Web & Product Attrs Match & \textcolor{blue}{Windows} \\ 
        \texttt{WebArena}~\cite{zhou2023webarena} & 241 & HTML & Web & url/text-match & \textcolor{blue}{Windows} \\
        
        \texttt{VisualWebArena}~\cite{koh2024visualwebarena} & 314 & HTML & Web & url/text/image-match & \textcolor{blue}{Windows} \\
        \texttt{Ferret-UI}~\cite{you2024ferretuigroundedmobileui} & 314 & HTML & Web & url/text/image-match & \textcolor{blue}{Windows} \\
        \texttt{OSWorld}~\cite{xie2024osworld} & 369 & - & Web & Device/Cloud state & \textcolor{BrownYellow}{Linux} \\
        
        \bottomrule
        \end{tabular}
    }
    \caption{Comparison of various platforms based on parallelization, templates, tasks per template, rewards, and supported platforms.}
    \vspace{-3mm}
    \label{tab:plf_cmp}
\end{table*}

Additionally, OmniACT \cite{kapoor2024omniact} offers a comprehensive platform for evaluating task automation across various desktop applications and natural language tasks. WebVoyager \cite{he2024webvoyager} introduces an automated evaluation protocol using GPT-4V, capturing screenshots during navigation and achieving an 85.3\% agreement with human judgments. Furthermore, Widget Captioning \cite{li2020widget} sets a benchmark for improving UI accessibility and interaction by providing 162,859 human-annotated phrases that describe UI elements from multimodal inputs, paving the way for advancements in natural language generation tasks. Above all, leveraging a diverse set of system signals provides a more comprehensive and accurate assessment of an agent's performance \cite{xie2024osworld}.

\begin{table*}[ht]
    \centering
    \resizebox{1\textwidth}{!}{ 
        \begin{tabular}{lcccccc}
        \toprule
        \textbf{Method} & \textbf{Input Type} & \textbf{Training} &\textbf{Memory}  & \textbf{Task} & \textbf{Multi-agents} \\ 
        
        \midrule \textbf{\textit{Prompt-based Methods}} \\ 
        ReAct~\cite{yao2022react} & Text & None & \greencheck & Web & \redcross \\
         MM-Navigator~\cite{yan2023gpt} & Image\&Text & None & \redcross & Apps+Web & \redcross \\
        MindAct~\cite{deng2024mind2web} & Text & None & \greencheck & Apps+Web & \redcross \\
        OmniAct~\cite{kapoor2024omniact} & Text & None & \redcross & Apps+Web & \redcross \\
        \midrule \textbf{\textit{Training-based Methods}} \\
        VUT~\cite{li2021vut} & Image\&Text & Pre-trained & \redcross & Web & \redcross \\
            
        Spotlight~\cite{li2022spotlight} & Image\&Text & Pre-trained & \redcross & Web & \redcross \\
       
        ScreenAI~\cite{baechler2024screenai} & Image\&Text & Pre-trained & \redcross & Web & \redcross \\
    
        ScreenAgent~\cite{niu2024screenagent} & Image\&Text & Pre-trained & \greencheck & Web & \redcross \\
    
        SeeClick~\cite{cheng2024seeclick} & Image\&Text & Pre-trained & \redcross & Web & \redcross \\
        
        \bottomrule
        \end{tabular}
        }
    \caption{Comparison of Mobile Agents: A Detailed Overview of Input Types, Models, Training Methods, Memory Capabilities, Tasks, and Multi-agent Support. Web* means synthesized web data. }
    \label{tab:mobile_agent}
\end{table*}

On desktop platforms, research has focused on evaluating how well LLM-based agents utilize APIs and software tools to complete tasks such as file management and presentations~\cite{qin2023toolllm,guo2023pptc}. AgentBench~\cite{liu2023agentbench} offers a flexible, scalable framework for evaluating agent tasks, while PPTC Benchmark~\cite{guo2023pptc} targets the evaluation of LLM-based agents' performance in PowerPoint-related tasks.

\subsection{Comparison of GUI Agents Performance}

On the Android GUI benchmarks, GUI agents demonstrate competitive performance across both static and grounding tasks. In Table 4 (AITW static evaluation), CogAgent and SeeClick achieve strong overall accuracy, with CogAgent reaching 76.88\% and SeeClick 76.20\%, outperforming most non-GUI agents and indicating robustness in decision-making scenarios involving complex interfaces. Notably, these GUI agents integrate layout or multimodal features to enhance UI understanding, with CoCo-LLAMA and MobileVLM also achieving 70\%+ accuracy across most splits.

On the grounding task benchmark ScreenSpot (Table 5), GUI agents such as CogAgent, SeeClick, and UGround-V1 also perform competitively. UGround-V1 achieves 73.3\% average accuracy, while SeeClick reaches 53.4\%. CogAgent shows relatively lower grounding performance (47.4\%) compared to its strong action accuracy on AITW, indicating a possible gap in its grounding capabilities under screenshot-based settings.

Overall, GUI agents with layout-aware supervision or multimodal visual encoding consistently outperform zero-shot baselines such as Qwen2-VL and Qwen2.5-VL. Among all, UI-R1-3B, a rule-based reinforcement learning agent designed for GUI interaction, achieves the best average accuracy on ScreenSpot (83.3\%), demonstrating the effectiveness of combining structured prompts, fine-grained visual grounding, and layout reasoning in GUI environments.

\end{document}